\documentclass[letterpaper, 10 pt, conference]{ieeeconf}

\usepackage{amsmath}
\usepackage{amsfonts}
\usepackage{amssymb}

\usepackage{amsthm}
\usepackage[ruled,vlined]{algorithm2e}
\usepackage{mathtools}
\usepackage{tabularx}
\usepackage{graphicx}
\usepackage{subfigure}
\usepackage{enumerate}
\usepackage{textcomp}
\usepackage{float}
\usepackage{url}
\usepackage{verbatim}
\usepackage{cite}
\usepackage[linkcolor=black,citecolor=black,urlcolor=black,colorlinks=true]{hyperref}

\graphicspath{{figures/}}
\IEEEoverridecommandlockouts
\DeclareMathOperator*{\argmin}{arg\,min}

\newcommand{\tp}{^{\mathrm{T}}}

\newcommand{\df}[1]{\mathrm{d}{#1}}

\newcommand{\rbrac}[1]{({#1})}
\newcommand{\rBrac}[1]{\left({#1}\right)}

\newcommand{\cbrac}[1]{\{{#1}\}}
\newcommand{\cBrac}[1]{\left\{{#1}\right\}}

\newcommand{\norm}[1]{\Vert{#1}\Vert}
\newcommand{\Norm}[1]{\left\Vert{#1}\right\Vert}
\newcommand{\abs}[1]{\vert{#1}\vert}

\newtheorem{proposition}{Proposition}

\begin{document}

    \title{\LARGE\bf Generating Large-Scale Trajectories Efficiently using \\ Double Descriptions of Polynomials}

    \author{Zhepei Wang\textsuperscript{1,2,3}, Hongkai Ye\textsuperscript{1,2,3}, Chao Xu\textsuperscript{1,2}, and Fei Gao\textsuperscript{1,2}
        \thanks{This work was supported by National Natural Science Foundation of China under Grant 62003299.}
        \thanks{\textsuperscript{1}State Key Laboratory of Industrial Control Technology, College of Control Science and Engineering, Zhejiang University, Hangzhou 310027, China. {\tt\{wangzhepei, hkye, cxu, fgaoaa\}@zju.edu.cn}}
        \thanks{\textsuperscript{2}Huzhou Institute, Zhejiang University, Huzhou 313000, China.}
        \thanks{\textsuperscript{3}National Engineering Research Center for Industrial Automation (Ningbo Institute), Ningbo 315000, China.}
    }

    \maketitle
    \thispagestyle{empty}
    \pagestyle{empty}

\begin{abstract}
    For quadrotor trajectory planning, describing a polynomial trajectory through coefficients and end-derivatives both enjoy their own convenience in energy minimization. We name them double descriptions of polynomial trajectories. The transformation between them, causing most of the inefficiency and instability, is formally analyzed in this paper. Leveraging its analytic structure, we design a linear-complexity scheme for both jerk/snap minimization and parameter gradient evaluation, which possesses efficiency, stability, flexibility, and scalability. With the help of our scheme, generating an energy optimal (minimum snap) trajectory only costs 1 $\mu s$ per piece at the scale up to 1,000,000 pieces. Moreover, generating large-scale energy-time optimal trajectories is also accelerated by an order of magnitude against conventional methods.
\end{abstract}

\section{Introduction}
\label{sec:Introduction}
Smooth polynomial trajectories generated from minimizing jerk/snap are widely used in the navigation of quadrotors~\cite{Richter2013PolynomialTP,Burri2015RealtimeVM,Oleynikova2020OpenMPF}. Among these applications, the double descriptions of polynomial trajectory are frequently involved. One description, consisting of piece coefficients and piece times, is convenient for cost evaluation and trajectory configuration. Another description, consisting of piece end-derivatives and times, is convenient and stable for cost minimization~\cite{Bry2015AggressiveFO}.

Although these double descriptions offer an efficient and accurate way to obtain energy-optimal trajectories, the overhead and instability are often inevitable caused by numerical transformations between them~\cite{De2017NewNSS}. Besides, piece times are coupled into transformations. Without knowing its structure, directly optimizing times becomes hard or quite inefficient. In this situation, many perturbed energy-optimal trajectories are often generated to obtain gradient information~\cite{Mellinger2011MinimumST}, thus ruining the convenience from descriptions.

To overcome these drawbacks, we study the transformation between double descriptions. Its concrete structure and analytic expression are clearly provided, which is indeed a diffeomorphism. Leveraging its analytic form, we first design a scheme for linear-complexity jerk/snap minimization. Unnecessary computation on transformation is eliminated from this scheme, making its speed faster than many known schemes by at least an order of magnitude. Utilizing the smoothness, we also derive an analytic gradient for waypoints and times, which also enjoys minimal complexity. The exact gradient information makes it possible to directly optimize all parameters under complex constraints.

\begin{figure}[t]
    \centering
    \includegraphics[width=1.0\columnwidth]{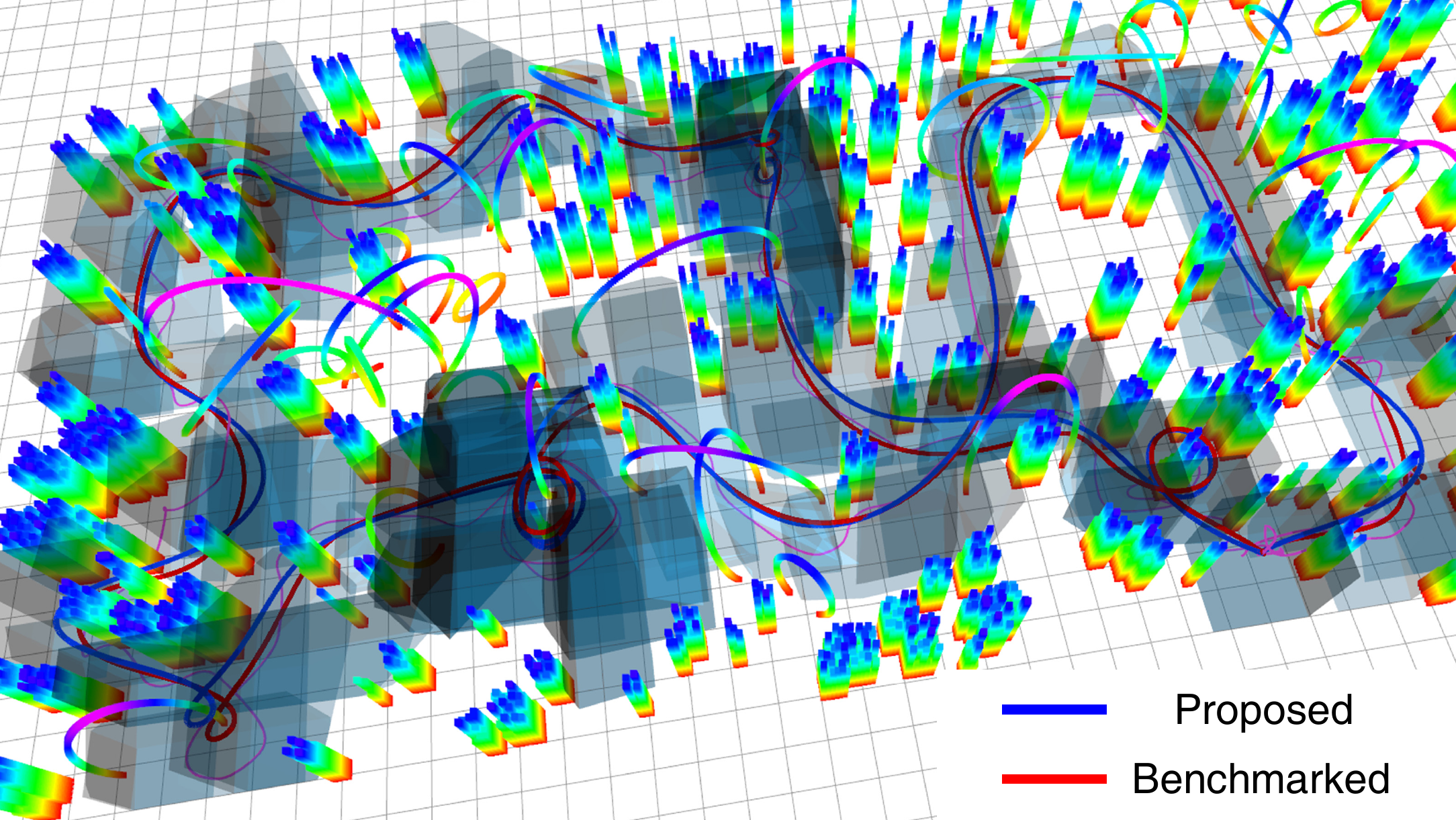}
    \caption{Large-scale energy-time optimal trajectories generated by the proposed scheme and the benchmarked scheme which is the global trajectory optimization in Teach-Repeat-Replan~\cite{Gao2020TeachRepeatReplanAC}. The proposed one takes $0.43$ seconds while the other takes $7.70$ seconds. The energy-time cost is $2643.43$  for the proposed one while it is $3269.76$ for the other. The efficiency is improved by an order of magnitude.\label{fig:LargeScaleSFC}}
    \vspace{-0.5cm}
\end{figure}

In this paper, we propose a framework based on the above results. Provided with a collection of waypoints and piece times, the minimization of jerk/snap is taken as a black box with promising efficiency. Through the cheap exact gradient, our framework directly optimizes intermediate waypoints and piece times to meet the safety constraints and dynamic limits, respectively. Its flexibility and efficiency are validated by applications and benchmarks on classic problems. Summarizing our contributions in this work:
\begin{itemize}
    \item An analytic transformation between double descriptions is derived.
    \item A linear-complexity minimum jerk/snap solution is designed with extreme efficiency.
    \item An analytic gradient for waypoints and piece times is provided with linear complexity.
    \item Applications and benchmarks on classic problems are provided to validate the superiority of our framework.
    \item High-performance implementation of solution and gradient computation are open-sourced for the reference of the community\footnote{Source code: \url{https://github.com/ZJU-FAST-Lab/large_scale_traj_optimizer}}.
\end{itemize}

\section{Related Work}
\label{sec:RelatedWork}
Quadrotor trajectory planning using polynomials has been prosperous since Mellinger et al.~\cite{Mellinger2011MinimumST}. They eliminate differential constraints from quadrotor dynamics via differential flatness. Then, enough smoothness of flat output trajectories guarantees the constraint satisfaction by default. It is thus quite different from common robotics trajectory generation where standard Nonlinear Programming (NLP) approaches must be employed to enforcing complicated dynamic constraints. Consequently, they conduct snap minimization for smooth flying trajectories with description of the first kind, where the coefficients are optimized through a Quadratic Programming (QP) with prescribed times and waypoints. The gradient for time is roughly estimated by solving perturbation problems. Richter et al.~\cite{Richter2013PolynomialTP} propose the description of the second kind which eliminates equality constraints in the QP, thus forming a closed-form solution. For large problems, its efficiency deteriorates because of the involved sparse matrix inverse. Safety constraints and dynamic limits are heuristically enforced, which can cause low trajectory quality. The two methods above provide many insights in quadrotor trajectory planning, even though there are weaknesses.

Many schemes are proposed to improve the above two frameworks. To improve efficiency, Burke et al.~\cite{Burke2020Generating} first propose a linear-complexity scheme to solve primal and dual variables of the QP, which generates trajectories with $500,000$ pieces in less than $3$ minutes. However, its efficiency is still not satisfactory because of the redundant problem size and block inverses. Optimizing waypoints and times is still not considered. Burri et al.~\cite{Burri2015RealtimeVM} and Oleynikova et al.~\cite{Oleynikova2020OpenMPF} directly optimize all end-derivatives of inner pieces through NLP, where piece times are fixed. Almeida et al.~\cite{Almeida2019RealTimeMS} train a supervised neural network to learn time allocation offline. Thus relatively good piece times can be allocated online. Our previous work~\cite{Wang2020AlternatingMTG} proposes an efficient optimization scheme for piece times, while the handling for safety constraints lacks flexibility.

Due to the seeming inflexibility in raw polynomial splines, other methods describe trajectories via control points of B-Splines and B\'ezier curves \cite{Preiss2017trajectorySCN,Gao2018OnlineSTG,Zhou2019RobustFAF}. The safety constraints and dynamic limits can be easily enforced via the convex hull property. Tordesillas et al.~\cite{Tordesillas2019Faster} utilize the property and optimize the interval allocation. A Mixed Integer Quadratic Programming (MIQP) is formulated to assign each trajectory piece into a convex region. However, the property can be conservative since the temporal profile of the trajectory is limited by the geometrical profile. To handle the issue, Gao et al.~\cite{Gao2020TeachRepeatReplanAC} propose a convex formulation to improve the dynamic performance of an already-known geometrical curve.

\section{Preliminaries}
\label{sec:Preliminaries}
Since the differential flatness of quadrotor has been validated in~\cite{Mellinger2011MinimumST}, polynomial trajectories are widely used in the continuous-time motion planning  for quadrotors. Consider an $(N+1)$-order $M$-piece spline whose $i$-th piece is indeed an $N$-degree polynomial $p_i(t)=c_i\tp\beta(t),~t\in\mathbf[0,T_i]$, where $c_i\in\mathbb{R}^{(N+1)}$ is a coefficient vector of this piece and $\beta(t)=\rbrac{1, t, t^2, \cdots, t^N}\tp$ a natural basis. The entire spline on the duration $[0,\tau_M]$ is defined by $p(t):=c_i\tp\beta( t-\tau_{i-1})$ where $t\in[\tau_{i-1},\tau_i]$ and $\tau_i=\sum_{j=1}^{i}{T_j}$.

For any given differentially flat quadrotor, high-order continuity is required to satisfy the differential constraints induced by the dynamics. Besides, the smoothness is also ensured by penalizing the integral of square of the high-order derivative. Assuming the order of the penalized derivative to be $s$, the cost function can be written as
\begin{equation}
J(c,T)=\int_{0}^{\tau_M}{p^{(s)}(t)^2}\df{t},
\end{equation}
where $c=(c_1\tp,\dots,c_M\tp)\tp\in\mathbb{R}^{M(N+1)}$ is a coefficient vector of the entire spline and $T=(T_1,\dots,T_M)\tp\in\mathbb{R}^M$ is a piece time vector. The cost $J$ implicitly decouples for each dimension~\cite{Bry2015AggressiveFO}, thus we only consider $1$-dimensional splines. According to the Linear Quadratic Minimum-Time (LQMT) problem~\cite{Verriest1991LinearQMT}, we set the degree as $N=2s-1$ hereafter because it is the optimal degree for the minimizer of $J$.

A spline $p(t)$ can be naturally described by the collection $\cbrac{c,T}$, i.e, the first description. Consider the minimization of $J(c,T)$ with fixed $T$ and some derivatives specified at certain timestamps. It can be formulated as a QP~\cite{Mellinger2011MinimumST} if $c$ is taken as a vector of decision variables. The second description is denoted by the collection $\cBrac{d,T}$ where $d\in\mathbb{R}^{2Ms}$ is an end-derivative vector $d=\rbrac{d_1\tp, \dots, d_M\tp}\tp$ where
\begin{equation}
\label{eq:EndpointDerivative}
d_i=\rbrac{p_i(0), \dots, p^{(s-1)}_i(0),p_i(T_i), \dots, p^{(s-1)}_i(T_i)}\tp.
\end{equation}
Bry et al.~\cite{Bry2015AggressiveFO} leverage the convenience of $\cBrac{d,T}$ to eliminate equality constraints in the QP so that a closed-form solution for the optimal $d$ is constructed. Obviously, the double descriptions of polynomial trajectories provide a lot of convenience for energy minimization.

\section{Efficient Solution and Gradient Computation With Double Descriptions}
In this section, we derive an explicit diffeomorphism between double descriptions of polynomial trajectories. Based on the transformation, we construct the solution for $J(c,T)$ with any $T$ in $O(M)$ linear complexity. Utilizing its properties, we obtain the analytical gradient for problem parameters, i.e., specified derivatives and $T$, which offers much flexibility to further improve the trajectory quality.

\subsection{Explicit Diffeomorphism Between Double Descriptions}
First, we explore the relation between parameterization spaces of both descriptions. We denote by $\mathbb{P}_\mathrm{c}:\mathbb{R}^{2Ms}\times\mathbb{R}_{>0}^{M}$ the space for $\cbrac{c,T}$ and by $\mathbb{P}_\mathrm{d}:\mathbb{R}^{2Ms}\times\mathbb{R}_{>0}^{M}$ the space for $\cBrac{d,T}$. The relation between $\mathbb{P}_\mathrm{c}$ and $\mathbb{P}_\mathrm{d}$ is given as below.
\begin{proposition}
    \label{ps:DiffeomorphismBetweenParameterizations}
    For a polynomial trajectory, denote by $\cbrac{c,T}$ its parameters in $\mathbb{P}_\mathrm{c}$ and by $\cBrac{d,T}$ its parameters in $\mathbb{P}_\mathrm{d}$. The map from $\cbrac{c,T}$ to $\cBrac{d,T}$ is a $C^{\infty}$-diffeomorphism between $\mathbb{P}_\mathrm{c}$ and $\mathbb{P}_\mathrm{d}$. An explicit diffeomorphism consists of an identity map on $T$ and a smooth bijection between $c$ and $d$:
    \begin{equation}
    \label{eq:ParemeterTransformation}
    d=\mathbf{A}_\mathrm{F}(T)c,~c=\mathbf{A}_\mathrm{B}(T)d,
    \end{equation}
    where $ \mathbf{A}_\mathrm{F}(T)=\oplus_{k=1}^{M}{\mathbf{A}_f(T_k)}$, $\mathbf{A}_\mathrm{B}(T)=\oplus_{k=1}^{M}{\mathbf{A}_b(T_k)}$, and $\oplus$ is the direct sum stacking all diagonal sub-matrices. The forward and backward sub-matrices are
    \begin{equation}
    \label{eq:ForwardBackwardMapMatrix}
    \mathbf{A}_f(t)=\begin{pmatrix}\mathbf{E}&\mathbf{0}\\ \mathbf{F}\rbrac{t}&\mathbf{G}\rbrac{t}\end{pmatrix},~\mathbf{A}_b(t)=\begin{pmatrix}\mathbf{U}&\mathbf{0}\\\mathbf{V}\rbrac{t}&\mathbf{W}\rbrac{t}\end{pmatrix},
    \end{equation}
    which are partitioned into four blocks in $\mathbb{R}^{s\times{s}}$. The entry at the $i$-th row and $j$-th column of any block is defined by
    \begin{align}
    \label{eq:ForwardSubmatrixE}
    &\mathbf{E}_{ij}=
    \begin{cases}
    (i-1)! & \mathit{if}~i=j,\\
    0 & \mathit{if}~i\neq{j},
    \end{cases}\\
    \label{eq:ForwardSubmatrixF}
    &\mathbf{F}_{ij}\rbrac{t}=
    \begin{cases}
    (j-1)!/(j-i)!\cdot t^{j-i}& \mathit{if}~i\leq{j},\\
    0 & \mathit{if}~i>j,
    \end{cases}\\
    \label{eq:ForwardSubmatrixG}
    &\mathbf{G}_{ij}\rbrac{t}=\frac{(s+j-1)!}{(s+j-i)!}\cdot t^{s-i+j},\\
    \label{eq:ForwardSubmatrixU}
    &\mathbf{U}_{ij}=
    \begin{cases}
    1/(i-1)! & \mathit{if}~i=j,\\
    0 & \mathit{if}~i\neq{j},
    \end{cases}\\
    \label{eq:ForwardSubmatrixV}
    &\mathbf{V}_{ij}\rbrac{t}=\frac{\sum_{k=0}^{s-\max{\rbrac{i,j}}}{(-1)^k\binom{s}{i+k}\binom{2s-j-k-1}{s-1}}}{(j-1)!~(-1)^i\cdot t^{s+i-j}},\\
    \label{eq:ForwardSubmatrixW}
    &\mathbf{W}_{ij}\rbrac{t}=\frac{\sum_{k=0}^{s-\max{\rbrac{i,j}}}{\binom{s-k-1}{i-1}\binom{2s-j-k-1}{s-1}}}{(j-1)!~(-1)^{i+j}\cdot t^{s+i-j}}.
    \end{align}
\end{proposition}
\begin{proof}
    The forward sub-matrix $\mathbf{A}_f(t)$ given in (\ref{eq:ForwardBackwardMapMatrix}), (\ref{eq:ForwardSubmatrixE}), (\ref{eq:ForwardSubmatrixF}), and (\ref{eq:ForwardSubmatrixG}), comes from the definition of $d_i$ in (\ref{eq:EndpointDerivative}). Actually, $\mathbf{A}_f(t)$ is a \textit{general confluent Vandermonde Matrix} generated from two variables $\lambda_0=0$ and $\lambda_1=t$, both with multiplicity $s$. According to \textit{Spitzbart's Theorem}~\cite{Schappelle1972InverseCVM}, the inverse of $\mathbf{A}_f(t)$ always exists for $\lambda_0\neq\lambda_1$, whose entries are exactly coefficients of a set of polynomials constructed from $\lambda_0$ and $\lambda_1$. The backward sub-matrix $\mathbf{A}_b(t)$ given in (\ref{eq:ForwardBackwardMapMatrix}), (\ref{eq:ForwardSubmatrixU}), (\ref{eq:ForwardSubmatrixV}), and (\ref{eq:ForwardSubmatrixW}), is derived following~\cite{Schappelle1972InverseCVM}. The process only involves lengthy but mechanical derivation thus is omitted here for brevity. Obviously, the map from $c_i$ to $d_i$ and the inverse are always smooth at any $T\in\mathbb{R}^M_{>0}$. The bijectivity and the smoothness imply a diffeomorphism.
\end{proof}
Proposition~\ref{ps:DiffeomorphismBetweenParameterizations} gives analytic transformations between double descriptions, which have long been evaluated unwisely. In~\cite{Richter2013PolynomialTP}, $\mathbf{A}_b(t)$ is numerically computed by inverting $\mathbf{A}_f(t)$ for any given $t$. In~\cite{Burri2015RealtimeVM}, the structure of $\mathbf{A}_f(t)$ is exploited so that $\mathbf{A}_b(t)$ is evaluated efficiently and stably via \textit{Schur complement}, where only the inverse of a sub-matrix is needed. Our previous work~\cite{Wang2020AlternatingMTG} has the fewest online computations, where coefficients in $\mathbf{A}_b(t)$ is offline numerically computed by setting $t=1$. However, all of these schemes involve numerically unstable matrix inverse. Proposition~\ref{ps:DiffeomorphismBetweenParameterizations} is free of these drawbacks for the exact analytic expression. The diffeomorphism structure can be further utilized to greatly improve the efficiency and quality of trajectory generation.

\subsection{Linear-Complexity Trajectory Generation}
Consider the following trajectory generation problem,
\begin{align}
\label{eq:MinimumControlEffort}
\min_{c,T}&~{J(c,T)}=\int_{0}^{\tau_M}{p^{(s)}(t)^2}\mathrm{d}t,\\
s.t.~&~p^{(j)}(0)=d_{0,j},~0\leq j<s,\\
&~p^{(j)}(\tau_M)=d_{M,j},~0\leq j<s,\\
&~p(\tau_i)=q_i,~0\leq i<M.
\end{align}
The initial and terminal derivatives are specified by $d_0=(d_{0,0},\dots,d_{0,s-1})\tp$ and $d_M=(d_{M,0},\dots,d_{M,s-1})\tp$, respectively. Each $q_i$ is a specified intermediate waypoint at $\tau_i$. The $s-1$ times continuous differentiability of $p(t)$ on $[0,\tau_M]$ is implicitly required by the definition of $J(c,T)$, which is not explicitly formulated as equality constraints here.

Although a closed-form solution of (\ref{eq:MinimumControlEffort}) is given by Bry et al.~\cite{Bry2015AggressiveFO}, its efficiency is limited by the numerical evaluation of $\mathbf{A}_b(t)$ and the sparse permutation matrix inverse. To attain the linear complexity, Burke et al.~\cite{Burke2020Generating} leverage the problem structure to calculate both primal and dual variables through a block tridiagonal linear equation system. However, it still needs frequently inverting sub-blocks. The size of the system is also redundant because of the dual variables. Therefore, we give a linear-complexity scheme with minimal problem size, where the matrix inverse is totally eliminated.

Consider the $\cBrac{d,T}$ description of $p(t)$. We only need to obtain all unspecified entries in $d$, which are indeed $p^{(j)}(\tau_i)$ for $1\leq i<M$ and $1\leq j<s$. Rewrite $d$ as
\begin{equation}
\label{eq:ParameterDecomposition}
d=\mathbf{P}(\bar{d}+\mathbf{B}\widetilde{d})
\end{equation}
where $\widetilde{d}\in\mathbb{R}^{(M-1)(s-1)}$ is a vector containing all unspecified entries. The constant vector $\bar{d}\in\mathbb{R}^{(M+1)s}$ is defined as
\begin{equation}
\label{eq:SpecifiedDerivatives}
\bar{d}=\rbrac{d_0\tp,q_1,0_{s-1},\dots,q_{M-1},0_{s-1},d_M\tp}\tp.
\end{equation}
where $0_{s-1}\in\mathbb{R}^{s-1}$ is a zero vector. The permutation matrix $\mathbf{P}=\rbrac{\mathbf{P}_1\tp,\dots,\mathbf{P}_M\tp}\tp$ is defined as
\begin{equation}
\mathbf{P}_i=\rBrac{\mathbf{0}_{2s\times(i-1)s},\mathbf{I}_{2s},\mathbf{0}_{2s\times(M-i)s}}.
\end{equation}
The matrix $\mathbf{B}=\rbrac{\mathbf{B}_1,\dots,\mathbf{B}_{M-1}}$ is defined as
\begin{equation}
\mathbf{B}_i=\rBrac{\mathbf{0}_{(s-1)\times is},\mathbf{D},\mathbf{0}_{(s-1)\times(M-i)s}}\tp,
\end{equation}
where $\mathbf{D}=\rbrac{0_{s-1},\mathbf{I}_{s-1}}$. It should be noted that computing (\ref{eq:ParameterDecomposition}) only involves orderly accessing entries. We do not really need to compute the matrix product considering that $\mathbf{P}$ and $\mathbf{B}$ is highly structured.

The cost function $J(c,T)$ indeed takes a quadratic form
\begin{equation}
\label{eq:CostFunctionQuadraticForm}
J(c,T)=c\tp\mathbf{Q}_\Sigma(T)c
\end{equation}
in which $\mathbf{Q}_\Sigma(T)=\oplus_{i=1}^{M}{\mathbf{Q}(T_i)}$. Note that the symmetric matrix $\mathbf{Q}(t)$ has an analytical form whose entries are simple power functions of $t$.  Its analytical form is provided by Bry et al. in the appendix of~\cite{Bry2015AggressiveFO}, to which we refer for details. Substituting (\ref{eq:ParemeterTransformation}) and (\ref{eq:ParameterDecomposition}) into (\ref{eq:CostFunctionQuadraticForm}) gives
\begin{equation}
J=(\bar{d}+\mathbf{B}\widetilde{d})\tp\mathbf{P}\tp\mathbf{H}_\Sigma(T)\mathbf{P}(\bar{d}+\mathbf{B}\widetilde{d})
\end{equation}
in which $\mathbf{H}_\Sigma(T)=\oplus_{i=1}^{M}{\mathbf{H}(T_i)}$ is a symmetric matrix. All its diagonal blocks is fully determined by the matrix function
\begin{equation}
\label{eq:AnalyticalMatrixFunction}
\mathbf{H}(t)=\mathbf{A}_b\tp(t)\mathbf{Q}(t)\mathbf{A}_b(t).
\end{equation}
Obviously, the analytical form of $\mathbf{H}(t)$ can be easily derived for a fixed $s$ by combining our Proposition \ref{ps:DiffeomorphismBetweenParameterizations} and the result from Bry et al.~\cite{Bry2015AggressiveFO}. Therefore, we omit the parameter $T$ and denote $\mathbf{H}_\Sigma=\oplus_{i=1}^{M}{\mathbf{H}_i}$ hereafter because it is trivial to obtain all diagonal blocks for any given $T$ in linear time and space.

Differentiating $J$ w.r.t. $\widetilde{d}$ gives
\begin{equation}
{\df{J}}/{\df{\widetilde{d}}}=2\mathbf{B}\tp\mathbf{P}\tp\mathbf{H}_\Sigma\mathbf{P}(\mathbf{B}\widetilde{d}+\bar{d}).
\end{equation}
The optimal $\widetilde{d}$ satisfies $\norm{{\df{J}}/{\df{\widetilde{d}}}}=0$, i.e.,
\begin{equation}
\label{eq:BandedSystem}
\mathbf{M}\widetilde{d}=\bar{b}
\end{equation}
where $\mathbf{M}=\mathbf{B}\tp\mathbf{P}\tp\mathbf{H}_\Sigma\mathbf{P}\mathbf{B}$ and $\bar{b}=-\mathbf{B}\tp\mathbf{P}\tp\mathbf{H}_\Sigma\mathbf{P}\bar{d}$.

Actually, the computation for $\mathbf{M}$ and $\bar{b}$ is quite easy. We partition each $\mathbf{H}_i$ into four square blocks as
\begin{equation}
\mathbf{H}_i=\begin{pmatrix}\boldsymbol\Gamma_i&\boldsymbol\Lambda_i\\\boldsymbol\Phi_i&\boldsymbol\Omega_i\end{pmatrix}.
\end{equation}
Expanding $\mathbf{M}$ gives
\begin{equation}
\mathbf{M}=\begin{pmatrix}
\boldsymbol\alpha_1&\boldsymbol\beta_2&\mathbf{0}&\cdots&\mathbf{0}&\mathbf{0}\\
\boldsymbol\gamma_2&\boldsymbol\alpha_2&\boldsymbol\beta_3&\cdots&\mathbf{0}&\mathbf{0}\\
\mathbf{0}&\boldsymbol\gamma_3&\boldsymbol\alpha_3&\cdots&\mathbf{0}&\mathbf{0}\\
\vdots&\vdots&\vdots&\ddots&\vdots&\vdots\\
\mathbf{0}&\mathbf{0}&\mathbf{0}&\cdots&\boldsymbol\alpha_{M-2}&\boldsymbol\beta_{M-1}\\
\mathbf{0}&\mathbf{0}&\mathbf{0}&\cdots&\boldsymbol\gamma_{M-1}&\boldsymbol\alpha_{M-1}
\end{pmatrix}
\end{equation}
where
\begin{gather}
\label{eq:DiagonalBlocks}
\boldsymbol\alpha_i=\mathbf{D}(\boldsymbol\Omega_i+\boldsymbol\Gamma_{i+1})\mathbf{D}\tp,\\
\label{eq:SubdiagonalBlocks}
\boldsymbol\beta_i=\mathbf{D}\boldsymbol\Lambda_i\mathbf{D}\tp,~\boldsymbol\gamma_i=\mathbf{D}\boldsymbol\Phi_i\mathbf{D}\tp.
 \end{gather}
Similarly, we partition $\bar{d}$ as
\begin{equation}
\bar{d}=\rbrac{\kappa_0\tp,\kappa_1\tp,\dots,\kappa_{M-1}\tp,\kappa_M\tp}\tp
\end{equation}
where each $\kappa_i\in\mathbb{R}^s$ is a constant vector. Expanding $\bar{b}$ gives
\begin{equation}
\bar{b}=\rbrac{b_1\tp,b_2\tp,\dots,b_{M-2}\tp,b_{M-1}\tp}\tp
\end{equation}
in which the $i$-th part can be computed as
\begin{equation}
\label{eq:VectorRHS}
b_i=-\mathbf{D}\rBrac{\boldsymbol\Phi_i\kappa_{i-1}+(\boldsymbol\Omega_i+\boldsymbol\Gamma_{i+1})\kappa_i+\boldsymbol\Lambda_i\kappa_{i+1}}.
\end{equation}

Now we conclude the procedure to obtain the optimal trajectory. Firstly, compute each $\mathbf{H}_i$ for any given $T$ using the analytical function $\mathbf{H}(t)$ defined in (\ref{eq:AnalyticalMatrixFunction}). Secondly, compute $\bar{d}$ according to its definition (\ref{eq:SpecifiedDerivatives}) for the specified derivatives. Thirdly, compute nonzero entries in $\mathbf{M}$ and $\bar{b}$ according to (\ref{eq:DiagonalBlocks}), (\ref{eq:SubdiagonalBlocks}) and (\ref{eq:VectorRHS}). Fourthly, solve the linear equation system (\ref{eq:BandedSystem}) using \textit{Banded PLU Factorization}~\cite{Golub1983MatrixCMP}. Finally, recover the optimal coefficient vector $c$ through (\ref{eq:ParameterDecomposition}) and (\ref{eq:ParemeterTransformation}).

The above-concluded procedure only costs $O(M)$ linear time and space complexity. There is no numerical matrix inverse needed in the whole procedure because $\mathbf{A}_b$ is overcome by deriving its analytical form. Moreover, the linear equation system needed to be solved only involves necessary primal decision variables, thus achieving the minimal problem size.

\subsection{Parameter Gradient Computation in Double Descriptions}
Although the double descriptions make it possible to obtain the solution of (\ref{eq:MinimumControlEffort}) in a cheap way, the resultant trajectory quality is still determined by the parameter of the problem, i.e., intermediate waypoints $q=\rbrac{q_1,\dots,q_{M-1}}\tp$ and piece times $T$. Further optimization on $q$ and $T$ is needed to improve the trajectory quality while maintaining feasibility. Therefore, we utilize the diffeomorphism between double descriptions of the trajectory to derive the analytical gradient w.r.t. $q$ and $T$. The gradient helps to obtain optimal $q$ and $T$, which bridges the gap in many traditional trajectory planning methods such as \cite{Oleynikova2016ContinuousTO} and \cite{GAO2019FlyingPC}.

For any pair of $q$ and $T$, denote by $\widetilde{d}(q,T)$ the corresponding optimal $\widetilde{d}$, which satisfies
\begin{equation}
\label{eq:UnconstrainedControlMinimization}
\widetilde{d}(q,T)=\argmin_{\widetilde{d}}{J(\mathbf{A}_\mathrm{B}(T)\mathbf{P}(\bar{d}+\mathbf{B}\widetilde{d}),T)}.
\end{equation}
Then, the optimal $c$, denoted by $c(q,T)$, is computed as
\begin{equation}
c(q,T)=\mathbf{A}_\mathrm{B}(T)\mathbf{P}(\bar{d}+\mathbf{B}\widetilde{d}(q,T)).
\end{equation}
Define a new cost as $\hat{J}(q,T):=J(c(q,T),T)$. The gradient we concern is indeed for $\hat{J}(q,T)$, i.e., $\partial\hat{J}/\partial{q}$ and $\partial\hat{J}/\partial{T}$.

Without causing ambiguity, we temporarily omit the parameters in $\hat{J}(q,T)$, $J(c,T)$, $\widetilde{d}(q,T)$, $c(q,T)$, $\mathbf{A}_\mathrm{B}(T)$ and $\mathbf{A}_\mathrm{F}(T)$ for simplicity.  As for the cost function in (\ref{eq:UnconstrainedControlMinimization}), $\widetilde{d}$ is a stationary point, which means its gradient w.r.t. $\widetilde{d}$ satisfies $\norm{\rbrac{{\partial J}/{\partial c}}\tp\mathbf{A}_\mathrm{B}\mathbf{P}\mathbf{B}}=0$. Now we know that the gradient computation in either $\mathbb{P}_c$ or $\mathbb{P}_d$ possesses its convenience. In $\mathbb{P}_c$, the gradient $\partial J/\partial T_i$ and $\partial J/\partial c_i$ both have easy-to-derive analytic expressions. In $\mathbb{P}_d$, the gradient of the cost (\ref{eq:UnconstrainedControlMinimization}) by $\widetilde{d}$ is already zero. Thus, we first express $\partial\hat{J}/\partial{T}$ and $\partial\hat{J}/\partial{q}$ by $\partial J/\partial T_i$ and $\partial J/\partial c_i$. Then, the diffeomorphism in Proposition~\ref{ps:DiffeomorphismBetweenParameterizations} is utilized to transform them into $\mathbb{P}_d$ such that terms relevant to $\widetilde{d}$ can be eliminated.

As for the gradient of $\hat{J}$ w.r.t. $T_i$, we have
\begin{align}
\label{eq:GradientByT}
\frac{\partial \hat{J}}{\partial T_i}&=\frac{\partial J}{\partial T_i}+\rBrac{\frac{\partial J}{\partial c}}\tp\frac{\partial\mathbf{A}_\mathrm{B}}{\partial T_i}\mathbf{P}(\bar{d}+\mathbf{B}\widetilde{d})~~~~~~~\nonumber\\
&+\rBrac{\frac{\partial J}{\partial c}}\tp\mathbf{A}_\mathrm{B}\mathbf{P}\mathbf{B}\frac{\partial \widetilde{d}}{\partial T_i} \nonumber\\
&=\frac{\partial J}{\partial T_i}+\rBrac{\frac{\partial J}{\partial c}}\tp\frac{\partial\mathbf{A}_\mathrm{B}}{\partial T_i}\mathbf{A}_\mathrm{F}c \nonumber\\
&=\frac{\partial J}{\partial T_i}+\rBrac{\frac{\partial J}{\partial c_i}}\tp\dot{\mathbf{A}}_b(T_i)\mathbf{A}_f(T_i)c_i.
\end{align}
As for the gradient of $\hat{J}$ w.r.t. $q_i$, we have
\begin{align}
\label{eq:GradientByQ}
\frac{\partial \hat{J}}{\partial q_i}&=\rBrac{\frac{\partial J}{\partial c}}\tp\mathbf{A}_\mathrm{B}\mathbf{P}\rBrac{\frac{\partial\bar{d}}{\partial q_i}+\mathbf{B}\frac{\partial\widetilde{d}}{\partial q_i}} \nonumber\\
&=\rBrac{\frac{\partial J}{\partial c}}\tp\mathbf{A}_\mathrm{B}\mathbf{P}\frac{\partial\bar{d}}{\partial q_i} \nonumber\\
&=\sum_{k=0,1}{\rBrac{\frac{\partial J}{\partial c_{i+k}}}\tp\mathbf{A}_b(T_{i+k})e_{(1-k)s+1}}.
\end{align}
where $e_j$ is the $j$-th column vector of $\mathbf{I}_{2s}$. As for $\partial J/\partial T_i$ and $\partial J/\partial c_i$, their analytic expressions are clearly derived in the appendix of~\cite{Bry2015AggressiveFO}.

It is obvious that the gradient computations given in (\ref{eq:GradientByT}) and (\ref{eq:GradientByQ}) are both irrelevant to the piece number $M$. Thus, computing the gradient w.r.t. $q$ and $T$ only costs $O(M)$ linear time and space complexity. Besides, all involved formulas have smooth analytic forms for $T\in\mathbb{R}^M_{>0}$, which much increase the efficiency of gradient computation.

Aside from efficiency, our analytical gradient scheme enjoy advantages in numerical stability. Specifically, the major numerical issue comes from the structure of $\hat{J}\rbrac{q,T}$ on $T$. As analyzed in~\cite{Wang2020AlternatingMTG}, if any piece time goes to zero, the trajectory energy goes to infinity, thus making $\hat{J}$ behave like a barrier function. Consequently, any scheme that evaluates gradient indirectly suffers instability from bad accuracy because this requires unrealistically small step size for finite difference~\cite{Mellinger2011MinimumST} near the barrier. In comparison, ours is free from this issue because of the available accurate gradient.

\section{Applications}
\subsection{Fast Minimum Jerk/Snap Trajectory Generation}
One natural application of the previous linear-complexity solution scheme is that we can compute minimum-jerk/snap trajectory with extreme efficiency. Actually, there are already many reliable trajectory generation schemes for this topic. One thing that really matters is whether we can use a scheme as a black box without even worrying about its computation burden. We show that our scheme almost satisfies the requirements on this black box.

To demonstrate the efficiency of our scheme, we benchmark it with several conventional schemes, including the QP scheme by Mellinger et al.~\cite{Mellinger2011MinimumST}, the closed-form scheme by Bry et al.~\cite{Bry2015AggressiveFO} and the linear-complexity scheme by Burke et al.~\cite{Burke2020Generating}.  As for Mellinger's scheme, we use OSQP~\cite{Stellato2020OSQP} to solve the QP formed by linear conditions and quadratic cost. As for Bry's scheme, both the dense and sparse linear system solvers are implemented to calculate the closed-form solution. As for Burke's scheme, we implement an optimized version of our own for the sake of fairness. More specifically, it only costs several seconds to calculate a minimum-snap trajectory with $500,000$ pieces, while the one in their paper is reported to cost more than $2$ minutes~\cite{Burke2020Generating}. All schemes are implemented in C++11 without any explicit hardware acceleration.

\begin{figure}[t]
    \centering
    \includegraphics[width=1.0\columnwidth]{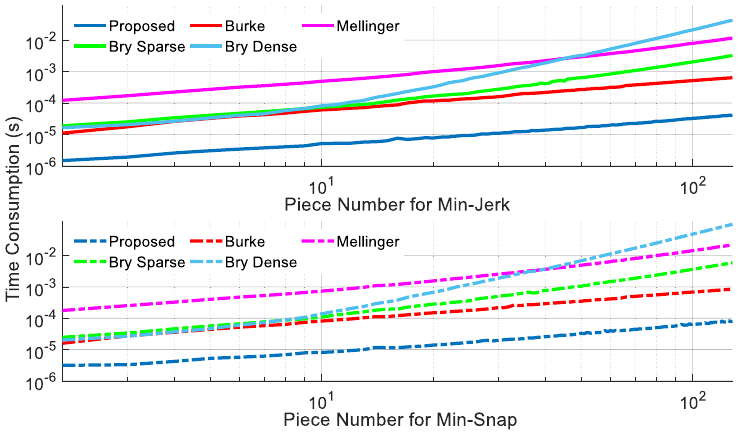}
    \caption{Benchmark of small-scale trajectory generation. Pink lines are from Mellinger's scheme~\cite{Mellinger2011MinimumST}. Lightblue lines are from Bry's scheme~\cite{Bry2015AggressiveFO}. Green lines are from its sparse version. Red lines are from Burke's scheme~\cite{Burke2020Generating}. Darkblue lines are from the proposed scheme. Solid lines are for jerk minimization. Dashed lines are for snap minimization.\label{fig:MinCtrlBenchmark1}}
    \vspace{-0.5cm}
\end{figure}

\begin{figure}[t]
    \centering
    \includegraphics[width=1.0\columnwidth]{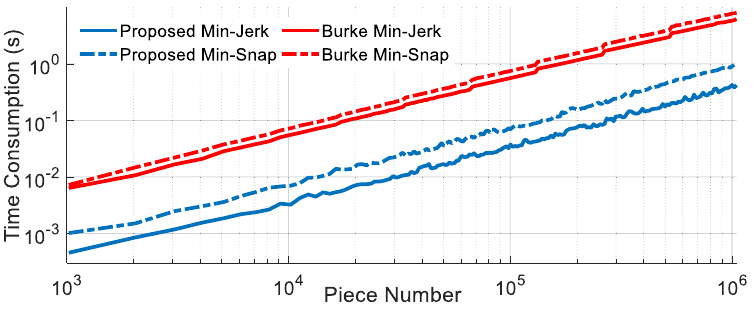}
    \caption{Benchmark of large-scale trajectory generation. Red lines are from Burke's scheme~\cite{Burke2020Generating}. Darkblue lines are from the proposed scheme. Solid lines are for jerk minimization. Dashed lines are for snap minimization.\label{fig:MinCtrlBenchmark2}}
    \vspace{-0.5cm}
\end{figure}

We benchmark all these five schemes on $3$-dimensional minimum jerk ($s=3$) and minimum snap ($s=4$) problems. All comparisons are conducted on an Intel Core i7-8700 CPU under Linux environment. For small-scale problems, the piece number ranges from $2$ to $2^7$. In each case, $100$ sub-problems are randomly generated to be solved by these schemes. The results are provided in Fig.~\ref{fig:MinCtrlBenchmark1}. For large-scale problems, we only benchmark two linear-complexity schemes, where the piece number is sparsely sampled from $2^{10}$ to $2^{20}$. The results are given in Fig.~\ref{fig:MinCtrlBenchmark2}.

As shown in Fig.~\ref{fig:MinCtrlBenchmark1}, Bry's closed-form solution and Burke's scheme are faster than Mellinger's scheme for small piece numbers. The dense version of Bry's scheme becomes the slowest since the cubic complexity for the dense solver dominates the time. Its sparse version still retains the efficiency as piece number grows. Due to the linear complexity, Burke's scheme and our scheme consume significantly less time than all other schemes. Moreover, ours is nearly an order of magnitude faster than Burke's at any problem scale in Fig.~\ref{fig:MinCtrlBenchmark1} or Fig.~\ref{fig:MinCtrlBenchmark2}. Intuitively, ours is able to generate trajectories with $10^6$ pieces in less than $1$ second.

\subsection{Fast Global Trajectory Optimization}
Now we give another application for the linear-complexity solution and gradient. We provide a simple example here to significantly improve the efficiency of large-scale global trajectory optimization.

A drawback in traditional minimum snap based schemes is the lack of flexibility to adjust the piece times and waypoints. Such kind of scheme can only obtain gradient information unwisely by solving several perturbed problems~\cite{Mellinger2011MinimumST}. To avoid this drawback, Gao et al.~\cite{Gao2020TeachRepeatReplanAC} propose a more flexible framework. It alternately optimizes the geometrical and temporal profile of a trajectory through two well-designed convex formulations. This framework generates high-quality large-scale trajectories within safe flight corridors while it cannot be done in real-time.

\begin{figure}[t]
    \centering
    \includegraphics[width=0.87\columnwidth]{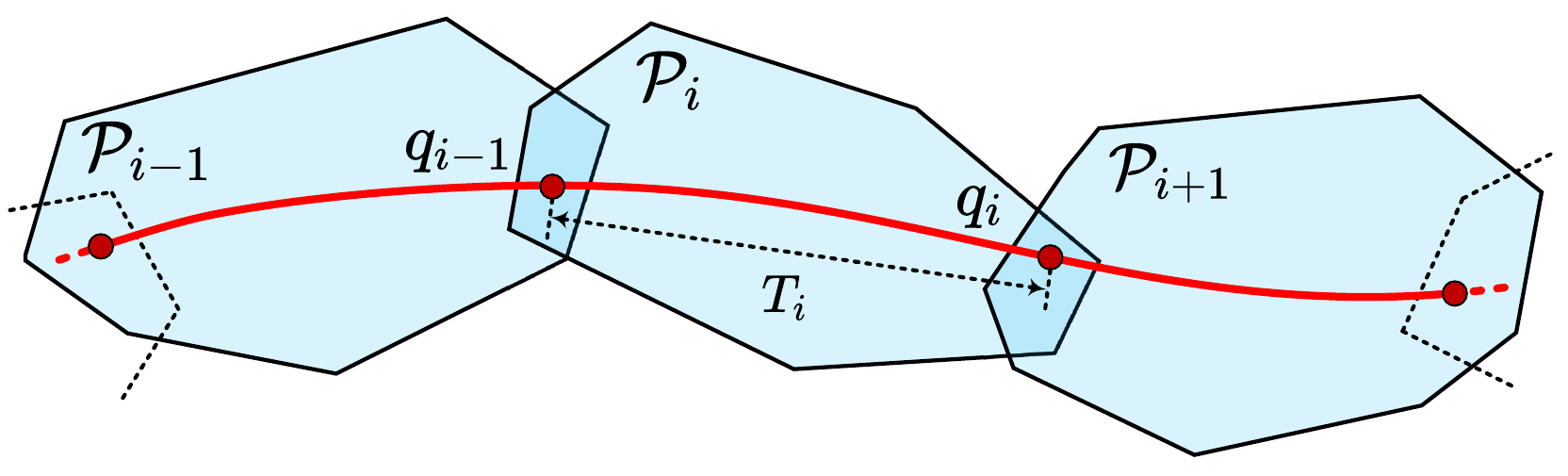}
    \caption{Each intermediate waypoint $q_i$ is confined within the intersection of two polyhedra $\mathcal{P}_{i}\cap\mathcal{P}_{i+1}$ using barrier functions. All piece times $T_i$ and intermediate waypoints $q_i$ are decision variables to be optimized. \label{fig:TrajSFC}}
    \vspace{-0.5cm}
\end{figure}

We assume that a polyhedron-shaped flight corridor has been generated as in~\cite{Gao2020TeachRepeatReplanAC}. The flight corridor $\mathcal{F}$ is defined as
\begin{equation}
\mathcal{F}=\bigcup_{i=1}^{M}{\mathcal{P}_i}
\end{equation}
where each $\mathcal{P}_i$ is a finite convex polyhedron
\begin{equation}
\mathcal{P}_i=\cBrac{x\in\mathbb{R}^3~\Big|~\mathbf{A}_ix\preceq b_i}.
\end{equation}
Besides, locally sequential connection is also assumed
\begin{equation}
\begin{cases}
\mathcal{P}_i\cap\mathcal{P}_j=\varnothing & \mathit{if}~\abs{i-j}=2,\\
\mathcal{P}_i\cap\mathcal{P}_j\neq\varnothing & \mathit{if}~\abs{i-j}\leq1.
\end{cases}
\end{equation}
For such a $\mathcal{F}$, the start and goal position is located in $\mathcal{P}_1$ and $\mathcal{P}_M$, respectively. As is shown in Fig.~\ref{fig:TrajSFC}, we assign each $3$-dimensional intermediate waypoint $q_i$ in the intersection $\mathcal{P}_i\cap\mathcal{P}_{i+1}$, which roughly ensures the trajectory safety.

To obtain the spatial-temporal optimal trajectory within $\mathcal{F}$, we optimize the following cost function
\begin{equation}
J_\Sigma(q,T)=\hat{J}(q,T)+J_F(q)+J_D(q,T).
\end{equation}
The cost term $\hat{J}(q,T)$ is also the $3$-dimensional version. The cost term $J_F(q,T)$ is just a logarithmic barrier term to ensure that each $q_i$ is confined within $\mathcal{P}_i\cap\mathcal{P}_{i+1}$, defined as
\begin{equation}
J_F(q)=-\kappa\sum_{i=1}^{M-1}{\sum_{j=i}^{i+1}\mathbf{1}\tp\ln{[b_j-\mathbf{A}_jq_i]}},
\end{equation}
where $\kappa$ is a constant barrier coefficient, $\mathbf{1}$ an all-ones vector with an appropriate length and $\ln{[\cdot]}$ the entry-wise natural logarithm. Actually, any $C^2$ clamped barrier function is an alternative to further eliminate the potential of the barrier in the interior of $\mathcal{P}_i\cap\mathcal{P}_{i+1}$, which can be easily constructed by following~\cite{Li2020IncrementalPC}. The cost term $J_D(q,T)$ is just a penalty to adjust the trajectory aggressiveness. It is defined as
\begin{align}
&J_D(q,T)=\rho_t \sum_{i=1}^{M}{T_i}+\\
&\rho_v \sum_{i=1}^{M-1}{g\rBrac{\Norm{\frac{q_{i+1}-q_{i-1}}{T_{i+1}+T_{i}}}^2-v_{m}^2}}+\nonumber\\
&\rho_a\sum_{i=1}^{M-1}{g\rBrac{\Norm{\frac{(q_{i+1}-q_{i})/T_{i+1}-(q_{i}-q_{i-1})/T_{i}}{(T_{i+1}+T_{i})/2}}^2-a_{m}^2}}\nonumber
\end{align}
where $g(x)=\max{\cbrac{x,0}}^3$ is a $C^2$ penalty function, $v_m$ the velocity limit and $a_m$ acceleration limit. The constant $\rho_t$ prevents the entire duration from growing too large. Constants $\rho_v$ and $\rho_a$ prevent the trajectory from being too aggressive. It also costs linear-complexity time for the value and gradient computation on $J_\Sigma(q,T)$. Then, we utilize the L-BFGS~\cite{Liu1989LBFGS} with strong Wolfe conditions as an efficient quasi-Newton method to minimize the cost function.

\begin{figure}[t]
    \centering
    \includegraphics[width=1.0\columnwidth]{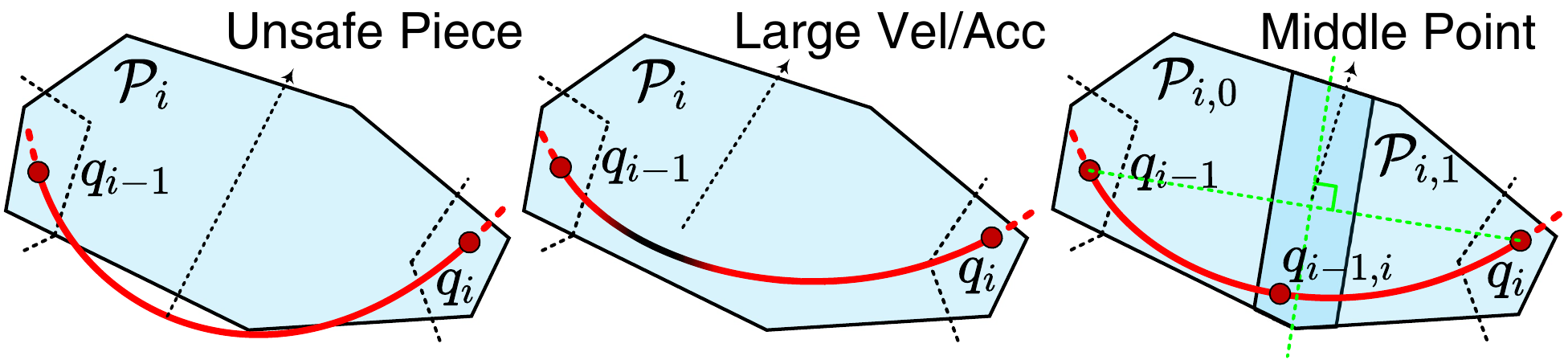}
    \caption{If constraints are much violated on the piece between $q_{i-1}$ and $q_i$, the corresponding polyhedron $\mathcal{P}_i$ is split into two intersecting polyhedra $\mathcal{P}_{i,0}$ and $\mathcal{P}_{i,1}$. where the barrier/penalty coefficients are also increased. Two perturbed planes near the perpendicular bisector of $q_{i-1}$ and $q_{i}$ are chosen as new facets. A new waypoint $q_{i-1,i}$ is also added in $\mathcal{P}_{i,0}\cap\mathcal{P}_{i,1}$.\label{fig:TrajSplitSFC}}
\end{figure}

As for constraints, it is possible that the interior part of a piece becomes unsafe or violates the limit too much. To handle such situations, we first utilize the feasibility checker proposed in~\cite{Wang2020AlternatingMTG} to locate such a piece. Then, the corresponding $\mathcal{P}_i$ is split into two intersecting parts, $\mathcal{P}_{i,0}$ and $\mathcal{P}_{i,1}$, as shown in Fig.~\ref{fig:TrajSplitSFC}. An intermediate waypoint $q_{i-1,i}$ is added as decision variables. Accordingly, we increase the corresponding penalty or barrier coefficient only for these two new pieces. In general, a larger penalty coefficient and shorter piece length make the soft constraint $J_D(q,T)$ tighter. A larger barrier coefficient in $J_F(q)$ makes $q_{i-1,i}$ more likely to be in the interior of $\mathcal{P}_{i,0}\cap\mathcal{P}_{i,1}$, which helps to ensure the safety. Empirically, these operations are seldom needed according to our simulations.

\begin{figure}[t]
    \centering
    \includegraphics[width=1.0\columnwidth]{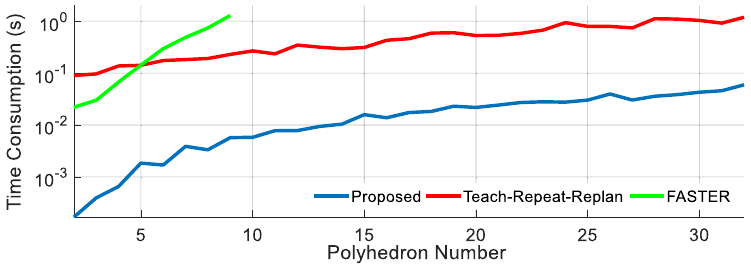}
    \caption{Benchmark of spatial-temporal trajectory optimization in safe flight corridor. The red line is from Teach-Repeat-Replan~\cite{Gao2020TeachRepeatReplanAC}. The green line is from FASTER~\cite{Tordesillas2019Faster}. The blue line is from the proposed one. FASTER only supports small-scale corridors since the computation time of its MIQP grows approximately in an exponential way.\label{fig:BenchmarkB}}
    \vspace{-0.5cm}
\end{figure}

We benchmark our simple scheme with the global trajectory optimizer in Teach-Repeat-Replan~\cite{Gao2020TeachRepeatReplanAC} which minimizes $J(c,T)+\rho_t\sum_{i=1}^{M}{T_i}$ under the same constraints. The interval allocation optimization in FASTER~\cite{Tordesillas2019Faster} is also compared here since it supports polyhedron-shaped corridor constraints. Due to the fact that it does not optimize time, we initialize it using total time from Teach-Repeat-Replan. We set $s=3$, $\rho_t=32.0$, $v_m=4.0$ and $a_m=5.0$ for all schemes and $\rho_v=\rho_a=128.0$ for the proposed one. The relative cost tolerance is set as $10^{-4}$ for ours and the default value for the other two open-source implementations. The time out is set as $1.5s$. For each size, the computation time is averaged over $10$ randomly generated corridors. The results from three methods is shown in Fig.~\ref{fig:BenchmarkB}. An intuitive comparison for large-scale trajectory generation is also provided in Fig.~\ref{fig:LargeScaleSFC} where a spatial-temporal optimal trajectory is generated by our scheme using significantly less time.

\section{Conclusion}
\label{sec:Conclusion}
In this paper, we explore and exploit the relation between double descriptions of quadrotor polynomial trajectory. The resultant linear-complexity solution and gradient computation provide much flexibility and efficiency in classic trajectory generation problems. Simple applications are provided to demonstrate their promising performance. Our future work focus on incorporating our results into existing local planners that use polynomial trajectories. It remains to be validated whether our results can greatly improve the trajectory quality in these planners without sacrificing the efficiency.

\newlength{\bibitemsep}\setlength{\bibitemsep}{0.00\baselineskip}
\newlength{\bibparskip}\setlength{\bibparskip}{0pt}
\let\oldthebibliography\thebibliography
\renewcommand\thebibliography[1]{
    \oldthebibliography{#1}
    \setlength{\parskip}{\bibitemsep}
    \setlength{\itemsep}{\bibparskip}
}
\bibliography{references}

\end{document}